# Can linguists better understand DNA?

Wang Liang

Huazhong University of Science and Technology, 430070, P.R. China

*To whom correspondence should be addressed. E-mail:wangliang.f@gmail.com

**[Abstract]**Multilingual transfer ability, which reflects how well models fine-tuned on one source language can be applied to other languages, has been well studied in multilingual pre-trained models. However, the existence of such capability transfer between natural language and gene sequences/languages remains under explored.This study addresses this gap by drawing inspiration from the sentence-pair classification task used for evaluating sentence similarity in natural language. We constructed two analogous tasks: DNA-pair classification(DNA sequence similarity) and DNA-protein-pair classification(gene coding determination). These tasks were designed to validate the transferability of capabilities from natural language to gene sequences. Even a small-scale pre-trained model like GPT-2-small, which was pre-trained on English, achieved an accuracy of 78% on the DNA-pair classification task after being fine-tuned on English sentence-pair classification data(XTREME PAWS-X). While training a BERT model on multilingual text, the precision reached 89%. On the more complex DNA-protein-pair classification task, however, the model's output was barely distinguishable from random output.Experimental validation has confirmed that the transfer of capabilities from natural language to biological language is unequivocally present. Building on this foundation, we have also investigated the impact of model parameter scale and pre-training on this capability transfer. We provide recommendations for facilitating the transfer of capabilities from natural language to genetic language,as well as new approaches for conducting biological research based on this capability.This study offers an intriguing new perspective on exploring the relationship between natural language and genetic language.

## 1 introduction

Advancements in large language models have transformed the landscape of artificial intelligence, with significant implications for bioinformatics. In the realm of nucleic acid analysis, specialized models like DNABert2, HyenaDNA, and ScBert have emerged to tackle classification and structural prediction challenges associated with DNA sequences (1~4). These models leverage the power of large language model to provide novel insights into genetic data.

Parallel developments have also occurred in protein-related research. Models such as ProTrans, ProteinBERT, and ESM2 have been introduced to excel in tasks like predicting protein structures and annotating their functions. These innovations highlight the versatility of large language models in addressing complex biological questions, bridging the gap between computational linguistics and molecular biology.

Furthermore, there are hybrid multi-modal large models that integrate natural language with DNA sequences, protein sequences, and structural data. Models such as BioMedGPT and InstructProtein are designed to address a variety of biological tasks. These models excel in applications like biological question answering, literature summarization, and predicting protein functions using natural language.

An interesting and important phenomenon associated with Multilingual Large Models (MLMs) is Language Capability Transfer. This phenomenon refers to the ability of a model, after being fine-tuned on one or more source languages, to effectively transfer the learned language understanding and generation capabilities to other unseen target languages. This not only demonstrates the model's strong generalization ability but also reveals the deep structures and patterns shared among different languages (14-17).

If such Language Capability Transfer also exists in gene-class large models, it would significantly expand the application scope of these gene models. Imagine if only supervised fine-tuning on English for tasks like summarization, question answering, and classification could then enable the model to handle DNA sequences for tasks such as summarization and classification. This would greatly enhance the efficiency of sequence search methods and other related processes.

**Challenges.** Due to the significant differences between natural language and gene language, evaluating and validating the transferability of capabilities from natural language to gene language presents a core challenge.

In natural language, typical evaluation datasets for cross-lingual pre-trained models include XTREME. XTREME encompasses four major categories—text classification, sequence labeling, sentence retrieval, and question answering—with nine sub-tasks in total. Models participating in XTREME are fine-tuned on English training data and then evaluated on test sets across 40 languages from 12 different language families. Although XTREME is not strictly parallel corpus-based, most multilingual data versions are still constructed based on English translations, with some data augmentation techniques such as word scrambling used to break one-to-one sentence correspondence while maintaining overall semantic alignment among languages (18).

If we aim to construct an evaluation dataset for transferring capabilities from English to DNA, it is essential to exclude tasks involving semantics, such as question answering and natural language inference, as the limited functional annotations of DNA sequences are insufficient to support building parallel corpora between English and DNA. Tasks like entity recognition and structure prediction also pose challenges because part-of-speech tagging in natural language and structural annotation in sequences differ significantly both quantitatively and qualitatively, making their construction difficult.

Among these tasks, sentence-pair similarity assessment is the least associated with semantics. While it evaluates sentence relationships, it focuses more on structural prediction capabilities. Therefore, we can draw inspiration from the PAWS-X dataset in XTREME, which is designed for sentence-pair similarity assessment, to construct a DNA sequence similarity judgment task dataset.

If a pre-trained model, after being fine-tuned on the English sentence similarity dataset from PAWS-X, can perform DNA sequence similarity judgments, this would indicate that natural language capabilities can be transferred to DNA language.

**Research Approach.**Our study draws on the PAWS-X dataset to construct two biosequence-related task datasets: DNA-pair classification for DNA sequence similarity judgment and DNA-protein-pair classification for gene-encoded protein similarity judgment. We then fine-tune GPT-2 and BERT series models on the PAWS-X English dataset and evaluate them on the aforementioned two biological sequence tasks. The results show that the precision of the DNA-pair classification task can reach up to 82%.

## 2 Materials and methods

### 2.1 Training datasets

Inspired by the PAWS-X evaluation dataset from XTREME, we constructed two biosequence similarity judgment datasets.

**PAWS-X Evaluation Dataset Overview**

PAWS-X is a sentence-pair similarity assessment task where each data entry consists of two sentences. A typical data entry looks like this:

```
{
  'sentence1': 'In Paris, in October 1560, he secretly met the English ambassador, Nicolas Throckmorton, asking him for a passport to return to England through Scotland.',
  'sentence2': 'In October 1560, he secretly met with the English ambassador, Nicolas Throckmorton, in Paris, and asked him for a passport to return to Scotland through England.',
  'label': 0
}
```

The task is to determine whether the two sentences express the same meaning, which is a binary classification problem. To extend this to other languages, the English sentence pairs were translated into six target languages (French, Spanish, German, Chinese, Japanese, and Korean) by professional translators and manually checked to ensure quality and consistency.

**DNA-Protein Pair Similarity Judgment Task**

Following a similar approach to PAWS-X, we first constructed a DNA-protein pair similarity judgment task based on gene coding rules. If a protein sequence corresponds to its DNA coding sequence, it is considered a positive example; otherwise, it is a negative example. The specific data design referenced the dataset used in Lucaone's paper. A typical data entry looks like this:

```
{
  'sentence1': 'ACCAGTGCTCAGGTTAACAAAATAATAAAAGGAA....',
  'sentence2': 'MASGRLQLLAFALALSGVSGVLAATLLPNWTVSVD...',
  'label': 0
}
```

The dataset contains approximately 25,600 entries, with a balanced number of positive and negative examples.

In Lucaone's DNA dataset, to increase the testing difficulty, each DNA sequence was extended to include an additional 100 nucleotides at both the 5' and 3' ends, preserving intron sequences within the data. We have removed these additional 100 base pairs (bp) of extra data.

### DNA-DNA/Protein-protein Sequence Pair Similarity Judgment Task

For the DNA-DNA sequence pair similarity judgment task, we used the DNA sequences from the DNA-protein task as source sequences. To construct similar sequences, we utilized NCBI BLAST to search public databases for similar sequences and generated pairwise alignment results. By setting strict thresholds (such as an E-value less than 1e-5), we filtered out highly credible homologous or similar sequence pairs.

For dissimilar sequences, we used genomic data from multiple model organisms, randomly selecting DNA sequences while ensuring similar lengths and low similarity scores (low local alignment scores). A typical data entry looks like this:

```
{
  'sentence1': 'ACCAGTGCTCAGGTTAACAAAATAATAAAAGGAA....',
  'sentence2': 'GGGCCGCCTTGCTGAAGCGGCATGGCCTCGGGG...',
  'label': 0
}
```

Construction of Protein-Protein Similar and Dissimilar Data Pairs and DNA-DNA Essential

Consistency.

## 2.2 Training Strategy

We used the GPT2-small model as a baseline for fine-tuning on the PAWS-X English dataset. Subsequently, we evaluated the model on two biosequence similarity judgment tasks: DNA-protein pair similarity judgment and DNA-DNA sequence pair similarity judgment.

Building on this foundation, we also tested other parameter scales of GPT2 models as well as BERT models.

For classification tasks, we utilized the default GPT2 model from Hugging Face. The model's output layer is configured with a commonly used softmax classification head; see Fig. 1 for specifics.

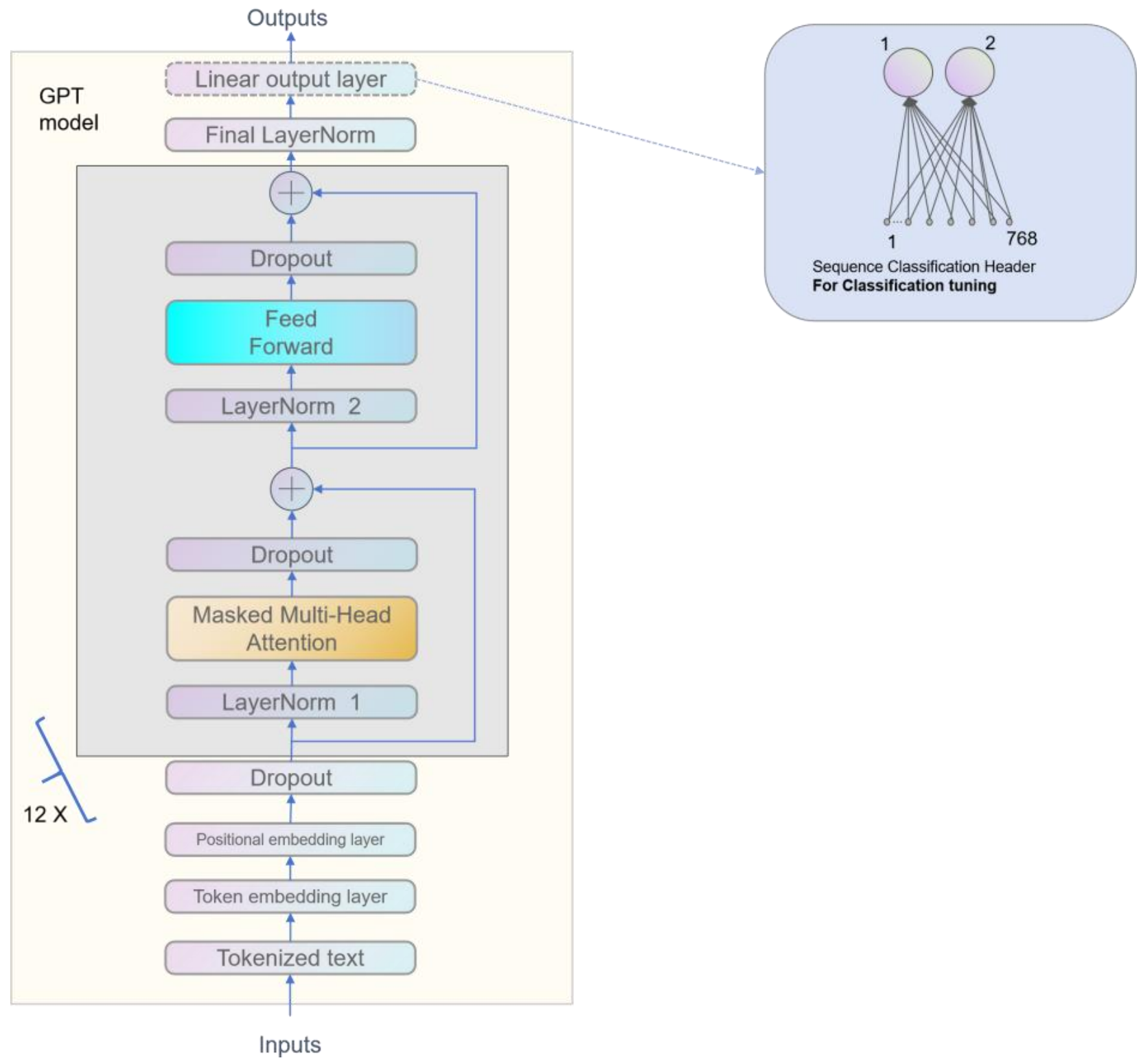

Fig.1 GPT2 model with sequence classification header

### Input and Output Format

To fine-tune GPT2, we concatenated the two sequences directly as the input string, with the label serving as the classification ID:

- **Input:** `sequence1 + sequence2`
- **Output:** `label`

**Note:** If using BERT-based models, the two sequences are connected using the `[SEP]` token. For GPT-based models, no separator is needed as the model itself determines the boundaries between the two sequences.

### Tokenization

Research on language capability transfer indicates that using a unified tokenizer is a necessary condition for transferring capabilities. Both the Byte Pair Encoding (BPE) tokenizer used in GPT2 models and the WordPiece tokenizer used in BERT models can process DNA and protein sequences. These tokenizers typically split sequences into tokens of one to three characters, which is a common method for feature extraction in biological sequences. Therefore, the choice of tokenizer does not affect our validation of language capability transfer.

**Fine-tuning Procedure**

1. Fine-tuning GPT2 on PAWS-X English Data:
   - We fine-tuned the GPT2-small model on the PAWS-X English dataset.
   - The input format was `sequence1 + sequence2`, and the output was the corresponding label.
2. Evaluation on Biosequence Similarity Tasks:
   - After fine-tuning, we evaluated the model on the DNA-protein pair similarity judgment task and the DNA-DNA sequence pair similarity judgment task.
3. Testing Other Model Variants:
   - We extended our experiments to include other parameter scales of GPT2 models and BERT models to compare performance.

**Tokenizer Considerations**

Given the importance of tokenization in language capability transfer, we ensured consistency by using:

- **GPT2 Models:** Utilized the BPE tokenizer.
- **BERT Models:** Used the WordPiece tokenizer with `[SEP]` tokens between sequences.

Both tokenizers effectively handle biological sequences, splitting them into short segments that are suitable for feature extraction in bioinformatics applications.

Due to the smaller scale of model parameters and the datasets size being in the tens of thousands of samples, we used the full-parameter fine-tuning method. Fine-tuning can be completed within about an hour on a single 4090 GPU.

# 3 Experimental Results

## 3.1 Basic results

As a binary classification task, we primarily used accuracy as the evaluation metric. Pre-trained models were fine-tuned on English datasets and then evaluated on gene-related tasks. For comparison, we also evaluated the models on French, German, and Chinese datasets.

For DNA sequences, we initially tested the raw DNA sequence pairs and found that the results were similar to random output. The main reason was that the each sentence in PAWS-X have an average string length of 113 characters, resulting in an average token length of 26 tokens after tokenization (approximately 4.5 characters per token) or 52 token for 2 concatenated sequences. In contrast, the corresponding DNA sequence had an average string length of 1336 characters, resulting in an average token length of 982 tokens after tokenization (approximately 1.92 characters per token) or 1964 token for 2 concatenated sequences, showing a significant difference. For foundational large models like GPT-2, the maximum supported input length is 1,024 tokens. However, the merged DNA sequences can reach a maximum length of 1,964 tokens, which necessitates truncating the DNA sequences and results in a loss of information.

Therefore, we truncated each DNA string to only include the first 50 base pairs (bp) and constructed similar or dissimilar DNA sequences. This way, the combined length of two DNA sequences became 80 to 100 bps. With an average of 1.92 characters per token, this resulted in a total of approximately 43 to 64 tokens, aligning closely with the PAWS-X datasets. Moreover, there is no information loss due to truncation.

The relevant test results are shown in the table below:

Table.1 Accuracy of DNA-pair classification(DNA sequence similarity)

| model | pretrain | finetune | test-en | test-fr | test-de | test-zh | test-dna | Test-DNA-Protein |
|---|---|---|---|---|---|---|---|---|
| gpt2 | en | en | 0.92 | 0.74 | 0.73 | 0.61 | 0.78 | 0.53 |
| bert | en | en | 0.91 | 0.77 | 0.73 | 0.52 | 0.84 | 0.49 |
| Bert-multi | Multi-lan | en | 0.94 | 0.86 | 0.83 | 0.77 | 0.89 | 0.52 |

We conducted experimental tests using three models: GPT-2 and BERT as baseline models, both trained primarily on English datasets with only a very limited amount of data from other languages. BERT-multi, on the other hand, was trained on over a hundred languages, predominantly English, but also including French, German, Chinese, and others.

These models were fine-tuned using the English similar sequence pairs from PAWS-X and then tested on similarity pair datasets in English, French, German, Chinese, and DNA sequences. The performance on natural language tasks was largely consistent with results reported in relevant literature.

In the DNA sequence similarity tests, the BERT model demonstrated significantly higher classification accuracy compared to GPT-2. Moreover, the multilingual BERT model outperformed the monolingual BERT model. This aligns with existing research findings that masked language models like BERT generally perform better in classification tasks than causal language models like GPT-2. A novel experimental discovery was that models trained on a greater variety of languages exhibited superior performance when transferred to DNA-related tasks.

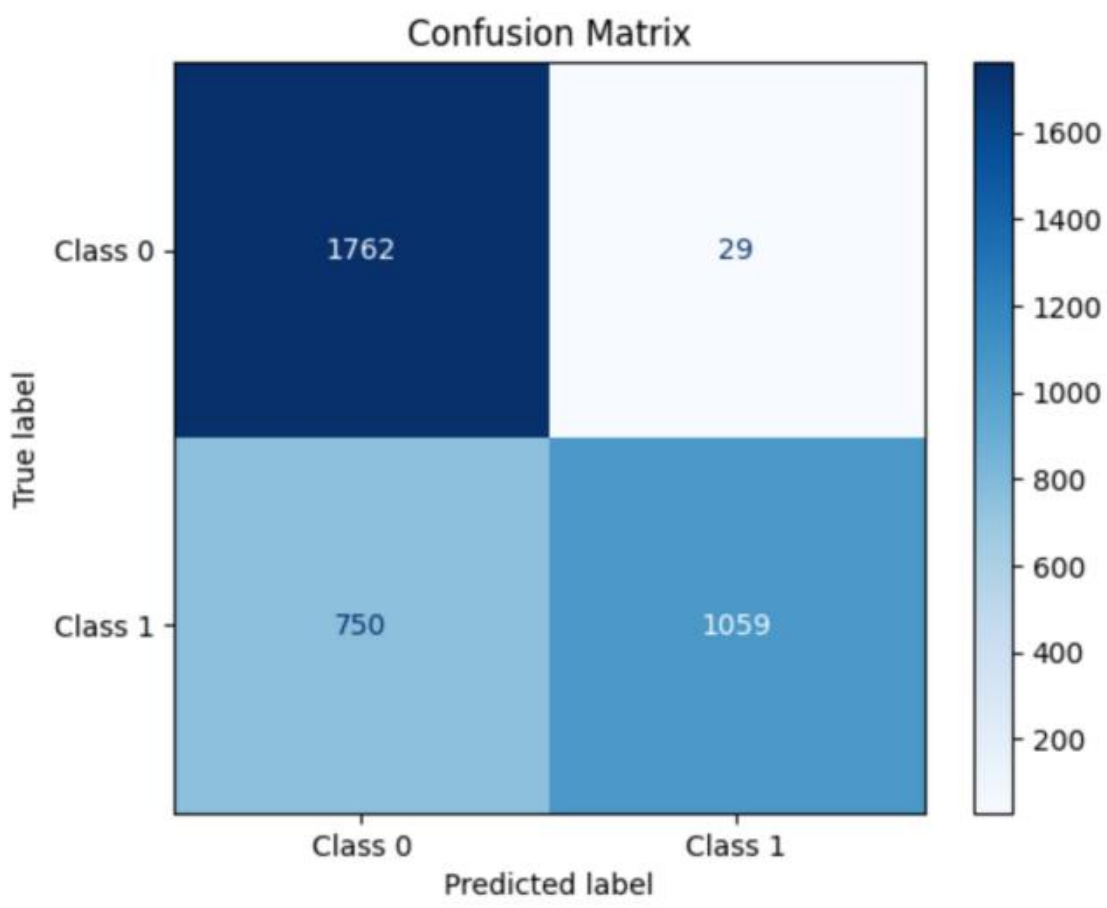

Fig.2 Confusion Matrix of DNA-pair classification, GPT2-small

Based on the test results of BERT-multi, we found that its classification accuracy on DNA similarity pairs was better than on Chinese and German. This is primarily because the construction of DNA similarity pairs was done using BLAST search without additional complex processing. In contrast, the similarity pairs in PAWS-X were specifically designed with semantic perturbations and other operations to increase the difficulty of the task.

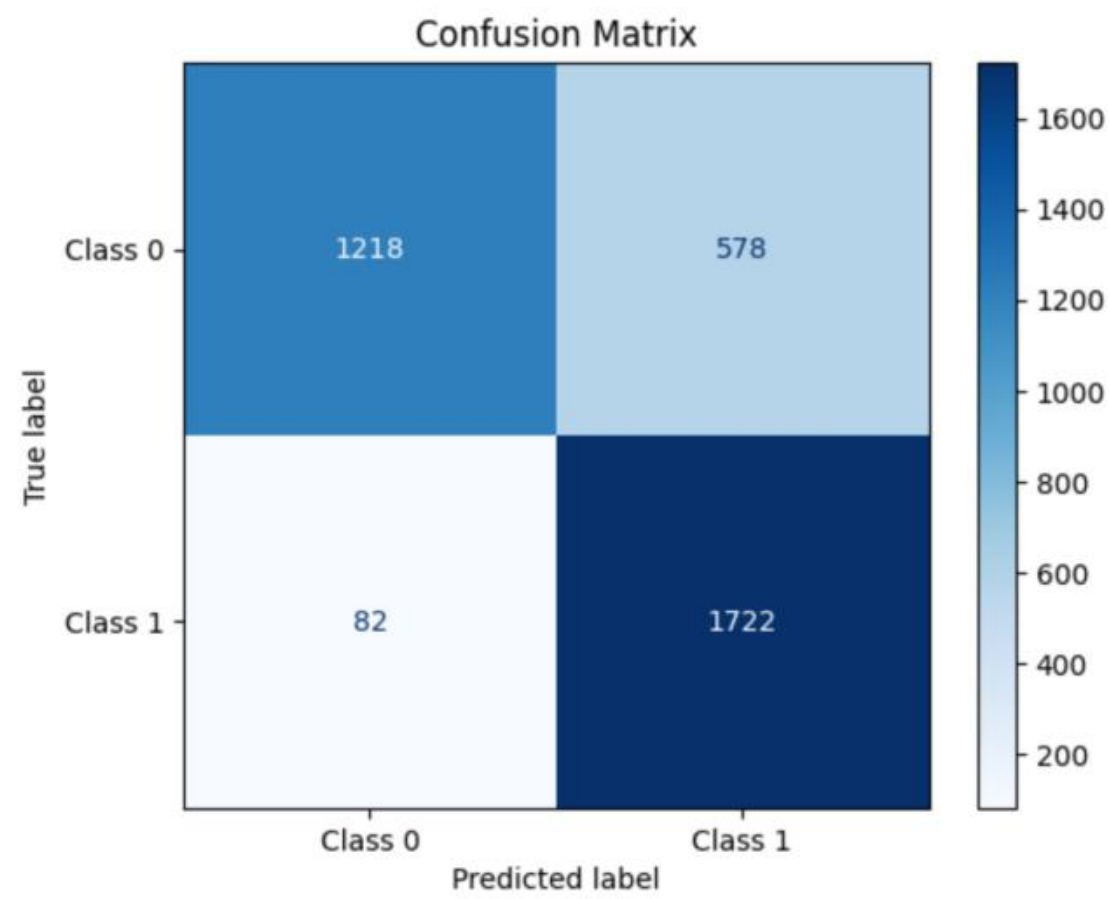

Fig.3 Confusion Matrix of DNA-pair classification, bert multilingual version

When the fine-tuned models were tested on DNA-protein pairs, their classification accuracy was close to 0.5, equivalent to random guessing. This suggests that the task of judging sequence similarity differs substantially from protein coding rules, making it challenging for capability transfer. To achieve effective transfer, it may be necessary to identify encoding tasks in natural language processing that are more analogous to those involved in protein coding. This could serve as an important direction for future research.

## 3.2 The Impact of Length

In the previous section's experiments, we discovered that sequence length has a clear impact on the transfer of language capabilities. Therefore, we further constructed biological sequences of varying lengths and imposed more stringent restrictions on the construction of similar sequences to increase the difficulty of the tests. The specific test datasets constructed are shown in the table below:

Table.2 single sequence length(BP) of different datasets

| Dataset | Mean length | Description |
| --- | --- | --- |
| Dna150s | 150 | Simple DNA Data Pairs. Among the similar pair data, 30% are completely identical pairs, and 40% have a containment relationship. |
| Dna150 | 150 | Similar Sequence Pairs: Pairs with a similarity score > 0.85 and a length difference of no more than 10%. These pairs do not include sequences that are completely identical or have a containment relationship.<br>Dissimilar Sequence Pairs: Pairs with a similarity score < 0.55 and a length difference of no more than 10%. |
| Dna50 | 50 | As above |
| Protein150 | 150 | As above. For reference, a 150 base pair (bp) DNA sequence corresponds to a protein sequence of approximately 50 amino acid residues in length. |
| Protein450 | 450 | As above. For reference, a 450 base pair (bp) DNA sequence corresponds to a protein sequence of approximately 150 amino acid residues in length. |
| Dna-protein | - | Lucaone paper origin data |

Because our datasets is designed for sequence similarity determination, each data point includes two sequences. The lengths listed in the table refer to the length of one sequence within each pair. The length distribution for the DNA150 datasets is illustrated in the figure below. The similarity scores for similar sequence pairs are based on the results from NCBI BLAST searches. In contrast, for dissimilar sequences, which are randomly selected, the similarity is assessed using local similarity measures.

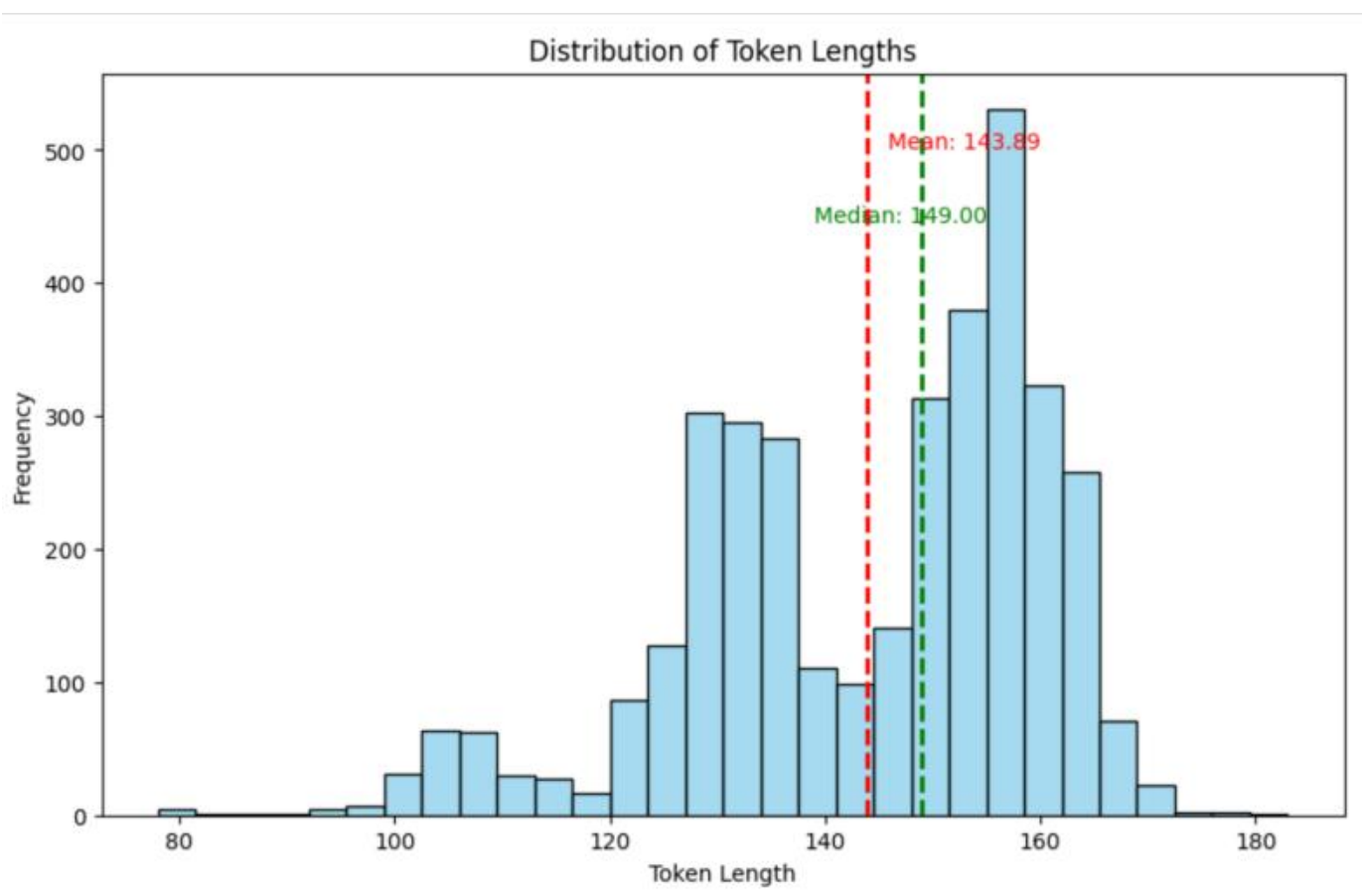

Fig4. Length distribution for the DNA150 datasets

Based on the aforementioned datasets, we conducted tests using both the GPT-2 and BERT-multi models. Considering the randomness inherent in model training, each experiment was repeated multiple times, and the best-performing model from each set of experiments was selected. The specific results are shown below:

Table.3 Accuracy of different datasets

| model | finetune | Dna150s | Dna150 | Dna50 | Protein150 | Protein450 | Dna-protein |
|---|---|---|---|---|---|---|---|
| GPT2 | en | 0.92 | 0.79 | 0.86 | 0.92 | 0.91 | 0.55 |
| Bert-multi | en | 0.9 | 0.94 | 0.91 | 0.96 | 0.93 | 0.46 |

The testing methodology remained consistent with the previous section. We began by performing classification fine-tuning on the English sequence similarity dataset from PAWS-X, followed by testing on other datasets. In the English data, the average sequence length is 50 characters, corresponding to approximately 11 tokens after tokenization. For the DNA50 dataset, sequences have an average length of 50 characters, resulting in about 26 tokens post-tokenization. The DNA150 dataset features sequences of 150 characters on average, leading to around 78 tokens. The DNA50 data closely resemble the English data in terms of sequence length. Protein data exhibit similar characteristics.

Our results showed that both models performed better on the Protein150 dataset compared to Protein450. In the tests on DNA50 and DNA150, the GPT-2 model achieved superior results on DNA50, whereas the BERT-multi model excelled on DNA150.

These findings suggest that when the length difference between source and target data is within one order of magnitude (up to 10 times), the impact on transfer performance is relatively minor. Generally, the closer the lengths of the source and target data, the better the transfer effect tends to be.

## 3.3 Training Stability

In the training of large models such as GPT-2, the choice of seed value can introduce variability in training outcomes. Our experiments demonstrated that when transferring English language capabilities to other natural languages, the influence of different seed settings is relatively minor. In contrast, when transferring these capabilities to DNA sequences, the impact of seed settings is considerably more significant.

Using the GPT-2 model as an example, we conducted fine-tuning on the English sequence similarity dataset from PAWS-X and subsequently tested the model's performance on French similarity pairs and DNA150 sequence similarity pairs. To assess the impact of seed settings, we randomly assigned different seed values and repeated the experiment multiple times, collecting statistics on the fine-tuning and testing results. The statistical outcomes for the French and DNA tests are illustrated in the figures below:

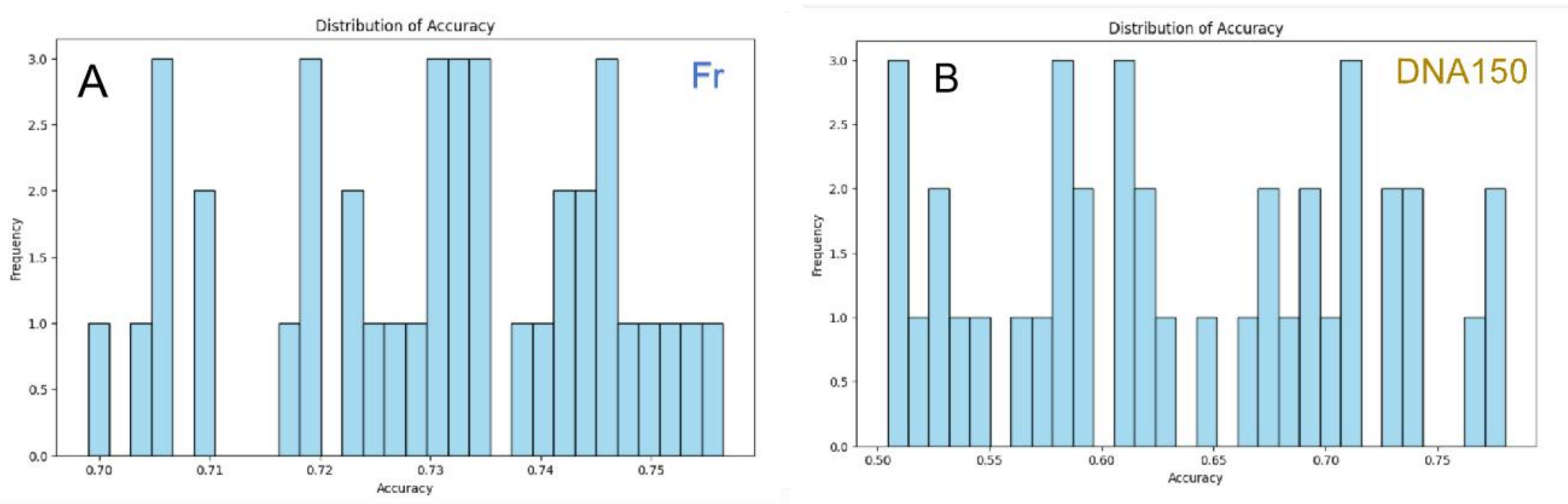

Fig5. Classification Accuracy Statistics. The left figure (A) shows the results for French, while the right figure (B) shows the results for DNA150 data.

From the French test results, we observe that the highest classification accuracy is 0.76, and the lowest is 0.70, with only a 6% difference between the best and worst performances. The statistical data are as follows: Mean: 0.730, Median: 0.732, Standard Deviation: 0.015.

In contrast, the DNA test results show a much greater variability. The highest classification accuracy is 0.79, while the lowest is only 0.52, which is close to random prediction. The difference between the best and worst results is substantial. The statistical data are as follows: Mean: 0.636, Median: 0.624, Standard Deviation: 0.082.

To more clearly illustrate the differences in language capability transfer between French and DNA sequences, we have plotted the results from multiple experiments on the same graph, as shown below:

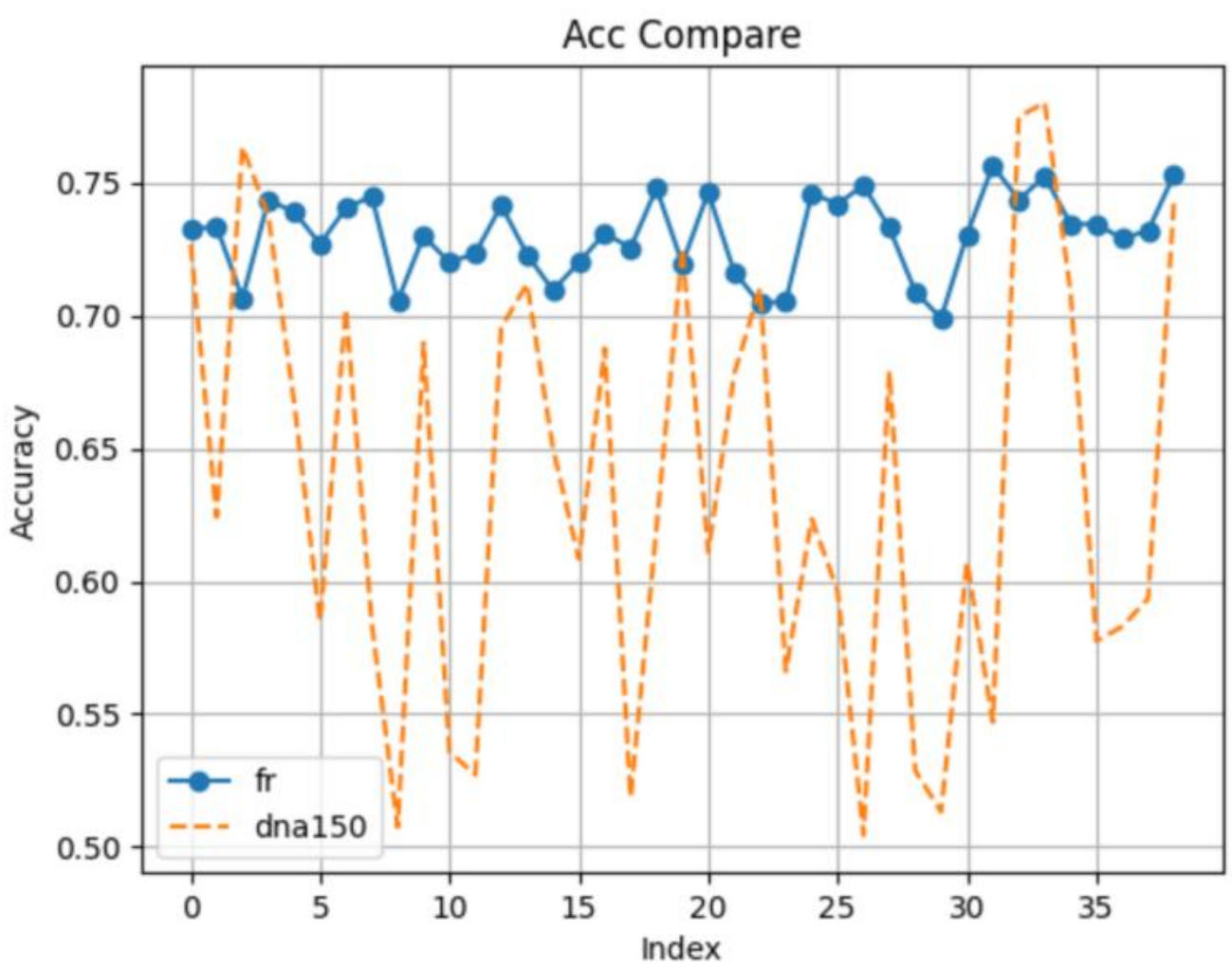

Fig.7. Results from Multiple Experiments for French and DNA150

In the figure, blue lines represent the test results for French, while orange lines represent the test results for DNA. The y-axis shows classification accuracy, and the x-axis indicates the index of each experiment. It can be observed that the test results for French exhibit much less variability, whereas the results for DNA sequences show significantly higher variability. However, there is a certain degree of synchronicity in the classification accuracy results between the two models.

## 3.4 Best Language

The PAWS-X test dataset includes not only English similarity text pairs but also similar text pairs in other languages such as French, German, and Chinese. We conducted further testing across all these languages. Specifically, we fine-tuned models using different natural language similarity sequences and then tested them on other languages, including DNA and protein sequences.

We present the test results for French and Chinese, as well as the results for various biological sequence pairs, as detailed below.

Table4: Accuracy of GPT2 model

| finetune | Test-fr | zh | Dna150s | Dna150 | Dna50 | Protein150 | Protein450 | Dna-protein |
|---|---|---|---|---|---|---|---|---|
| en | 0.76 | 0.64 | 0.92 | 0.79 | 0.86 | 0.92 | 0.91 | 0.55 |
| Fr | 0.85 | 0.67 | 0.88 | 0.78 | 0.66 | 0.89 | 0.92 | 0.54 |
| De | 0.83 | 0.68 | 0.9 | 0.82 | 0.7 | 0.85 | 0.91 | 0.57 |
| Es | 0.82 | 0.68 | 0.92 | 0.82 | 0.74 | 0.87 | 0.94 | 0.54 |
| Zh | 0.7 | 0.74 | 0.89 | 0.85 | 0.69 | 0.95 | 0.96 | 0.54 |
| Ja | 0.74 | 0.74 | 0.85 | 0.78 | 0.63 | 0.84 | 0.93 | 0.55 |
| Ko | 0.7 | 0.73 | 0.85 | 0.76 | 0.67 | 0.81 | 0.97 | 0.54 |

We used the GPT-2 model and fine-tuned it on similarity determination corpora from the PAWS-X dataset in multiple languages, including English (en) and French (fr), before testing it on similarity pair datasets in other languages. The evaluation metric used for testing was accuracy.

It can be observed that the model fine-tuned on the Chinese portion of the PAWS dataset performed relatively better on biological sequence tests.

Table5: Accuracy of bert Multi model

| finetune | Test-fr | zh | Dna150s | Dna150 | Dna50 | Protein150 | Protein450 | Dna-protein |
|---|---|---|---|---|---|---|---|---|
| en | 0.88 | 0.78 | 0.9 | 0.94 | 0.91 | 0.96 | 0.93 | 0.46 |
| Fr | 0.9 | 0.8 | 0.96 | 0.96 | 0.88 | 0.95 | 0.92 | 0.54 |
| De | 0.88 | 0.79 | 0.93 | 0.96 | 0.91 | 0.89 | 0.97 | 0.54 |
| Es | 0.89 | 0.8 | 0.9 | 0.95 | 0.93 | 0.96 | 0.99 | 0.46 |
| Zh | 0.85 | 0.83 | 0.91 | 0.96 | 0.92 | 0.99 | 0.99 | 0.53 |

| Ja | 0.84 | 0.82 | 0.88 | 0.91 | 0.9 | 0.97 | 0.99 | 0.55 |
|---|---|---|---|---|---|---|---|---|
| Ko | 0.84 | 0.81 | 0.73 | 0.84 | 0.89 | 0.97 | 0.98 | 0.54 |

We used the BERT-Multi model and fine-tuned it on similarity determination corpora from the PAWS-X dataset in multiple languages, including English (en) and French (fr), before testing it on similarity pair datasets in other languages. The evaluation metric used for testing was accuracy.

It can be observed that the model fine-tuned on the Chinese portion of the PAWS dataset performed relatively better on biological sequence tests.

## 3.5 The Impact of Pre-training

In both the GPT-2 and BERT-Multi models, pre-training was conducted using natural language data, which did not include biological sequences. To investigate the impact of pre-training on language transfer capabilities, we chose to train a GPT-2-small model from scratch.

**Custom Pre-training Setup**

Pure English GPT-2 Model (gpt2-small-openweb):We referenced the GPT-2 model architecture and used a subset of the OpenWeb English dataset (approximately 30 GB) to pre-train a pure English GPT-2-small model (gpt2-small-openweb) using around 10 GB of data.

Multimodal Large Model with Mixed Sequences (gene_eng_gpt2):Using the same 10 GB of English data, along with 10 GB of DNA sequences and 10 GB of protein sequences, we pre-trained a multimodal large model (gene_eng_gpt2) that incorporates various types of sequence data.

For both models, we constructed tokenizers using Byte Pair Encoding (BPE) and trained them from scratch.

Fine-tuning and Testing.After completing the pre-training phase, we fine-tuned both models using the PAWS-X English similarity determination dataset. Subsequently, we tested the models on other datasets. The test results are summarized as follows:

Table6. Accuracy of pretrained gpt2 model

| Model | pretrain | finetune | Test-fr | zh | Dna150s | Dna150 | Dna50 | Protein150 | Protein450 | Dna-protein |
|---|---|---|---|---|---|---|---|---|---|---|

| gpt2-small-openweb | en | en | 0.75 | 0.67 | 0.79 | 0.72 | 0.80 | 0.95 | 0.74 | 0.55 |
|---|---|---|---|---|---|---|---|---|---|---|
| gene_eng_gpt2 | en+DNA+protein | en | 0.72 | 0.67 | 0.85 | 0.84 | **0.67** | 0.96 | 0.96 | 0.55 |

As can be seen from the table above, the model pre-trained on English, DNA, and protein sequences demonstrates stronger language transfer capabilities compared to the model pre-trained solely on English. In the Dna50 data tests, due to the use of a model pre-trained on DNA data, the token lengths for word segmentation are very short, resulting in relatively poorer classification performance."

## 3.6 The Impact of Model Parameter Scale

Given the rapid expansion in the parameter scales of large models, we also tested the impact of parameter scale on multilingual capability transfer using different versions of the GPT-2 model. We fine-tuned models with varying parameter scales — ranging from smallest to largest: GPT-2-small, GPT-2-medium, GPT-2-large, and GPT-2-XL—on the PAWS-X English similarity pair determination dataset. Subsequently, we tested these models on other datasets. The test results are summarized as follows:

Table.7. Accuracy of different model

| Model | fr | zh | Dna150s | Dna 150 | Dna50 | Protein 150 | Protein450 | Dna-protein |
|---|---|---|---|---|---|---|---|---|
| gpt2-small | 0.76 | 0.64 | 0.92 | 0.79 | 0.86 | 0.92 | 0.91 | 0.55 |
| gpt2-medium | 0.81 | 0.66 | 0.84 | 0.6 | 0.72 | 0.89 | 0.62 | 0.5 |
| gpt2-large | 0.82 | 0.66 | 0.9 | 0.75 | 0.78 | 0.93 | 0.82 | 0.56 |
| gpt2-Xlarge | 0.83 | 0.67 | 0.92 | 0.81 | 0.88 | 0.93 | 0.92 | 0.52 |

As illustrated in the table above, the GPT-2-small model serves as a baseline. Despite having the smallest parameter scale, it demonstrates robust language transfer capabilities that are competitive with larger models.

GPT-2-small: Performs comparably well, serving as a strong baseline.

GPT-2-medium: Shows a slight decrease in performance compared to GPT-2-small.

GPT-2-large: Performance improves again, surpassing the medium-sized model.

GPT-2-XL: The largest model in terms of parameter scale, it achieves the best language transfer capability. However, the improvement over GPT-2-small is not particularly substantial.

This suggests that while increasing model size generally enhances performance, the gains may not always be proportional to the increase in parameters. Smaller models like GPT-2-small can still offer strong performance in language transfer tasks.

## 3.7 Word Vector Analysis

Currently, theoretical analyses of language transfer capabilities predominantly involve using parallel sentence pairs or word pairs from different languages. Large models generate corresponding word or sentence vectors for these pairs, and then analyze the distances or similarities between these vectors. A typical observation is that vector pairs generated by lower network layers tend to be farther apart, whereas those generated by higher layers are closer. Additionally, vector pairs fine-tuned with datasets like PAWX exhibit closer distances compared to those from pre-trained models without fine-tuning.

However, for natural language and DNA sequences, there aren't explicit parallel sentences or word pairs. Therefore, we adopted a flexible approach focusing on comparing the vocabularies of natural language (using English) and DNA. The specific steps are as follows:

1. Word Vector Generation: Using the entire set of English words and DNA vocabulary, we generate word vectors. The input is specific words, and we use the last hidden layer of the GPT-2 network, applying average pooling to obtain a vector representation for each word. Each word is represented as a 768-dimensional vector.

2. Measuring Similarity Between DNA Word Vectors and English Word Vectors: We use the Mean Nearest Neighbor Distance (MNND) method to calculate the similarity. This method computes the distance from each vector in one group to its nearest neighbor in the other group and then takes the average of these distances. This allows us to measure the "closeness" between the two groups of vectors. Specifically, we calculate the nearest distance from each DNA word vector to the group of English word vectors and then take the average.

Based on the aforementioned methods, we used the `gene_eng_gpt2` model as our pre-trained model, which was pre-trained on both English and DNA sequences. We calculated the distance between DNA word vectors and English word vectors using this model, resulting in a mean nearest neighbor distance of 50.09.

As a comparison, we fine-tuned the `gene_eng_gpt2` model using English paraphrase pairs from the PAWS dataset to obtain the fine-tuned model `gene_eng_gpt2_ft`. Using this fine-tuned model to generate word vectors, the distance between DNA word vectors and English word vectors decreased to 29.79.

This significant reduction in distance demonstrates that after fine-tuning, the overall proximity between DNA vocabulary and English vocabulary has substantially increased. This observation can help explain the phenomenon of natural language capabilities transferring to DNA language.

To visualize this more clearly, we employed PCA for dimensionality reduction and plotted the DNA word vectors and English word vectors in a two-dimensional graph. The specific visualization steps are as follows:

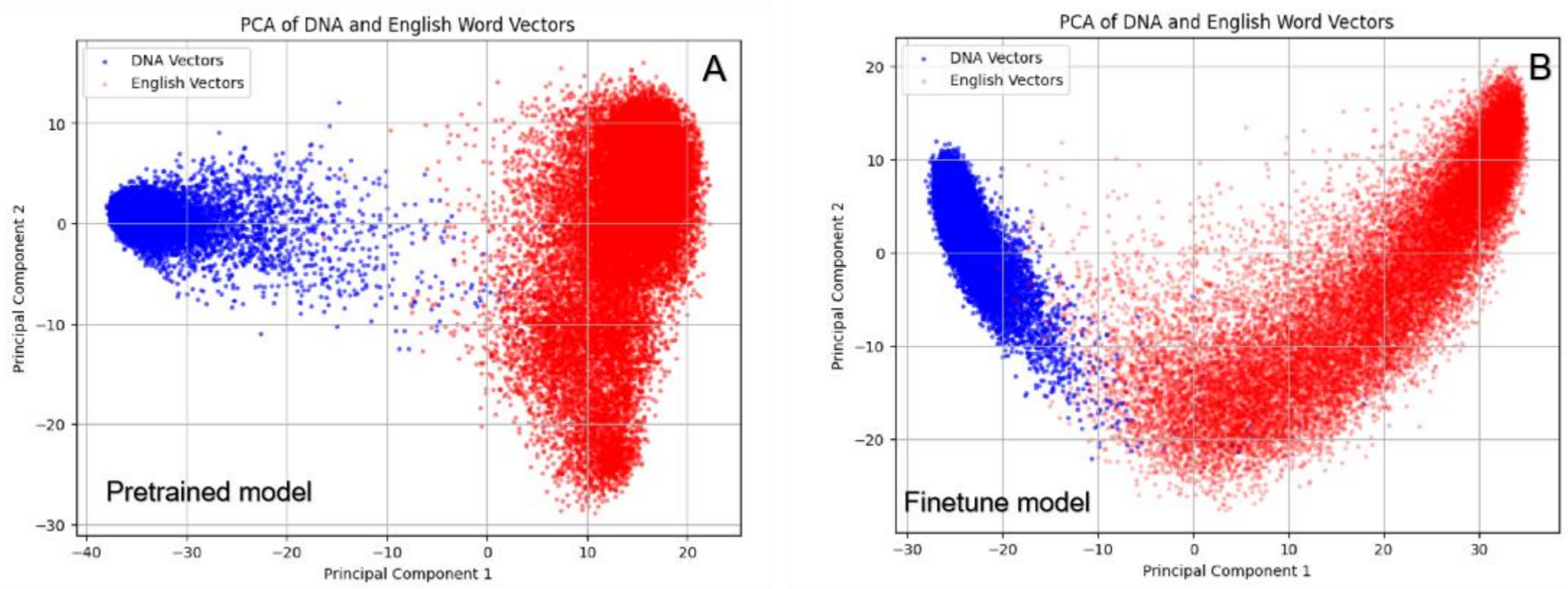

Fig.8.On the left, Figure A shows the word vectors generated by the pre-trained model. Blue points represent DNA word vectors, and red points represent English word vectors. On the right, Figure B shows the word vectors generated by the fine-tuned model. Again, blue points represent DNA word vectors, and red points represent English word vectors.In Figure A, it is evident that the DNA word vectors and English word vectors are generally far apart, with a clear boundary between them. In contrast, in Figure B, the DNA word vectors and English word vectors are noticeably closer, with some overlapping regions. This further illustrates the reason for language capability transfer in the fine-tuned model.

Furthermore, when we plot the word vectors for DNA, English, and proteins on a single graph, we can draw similar conclusions:

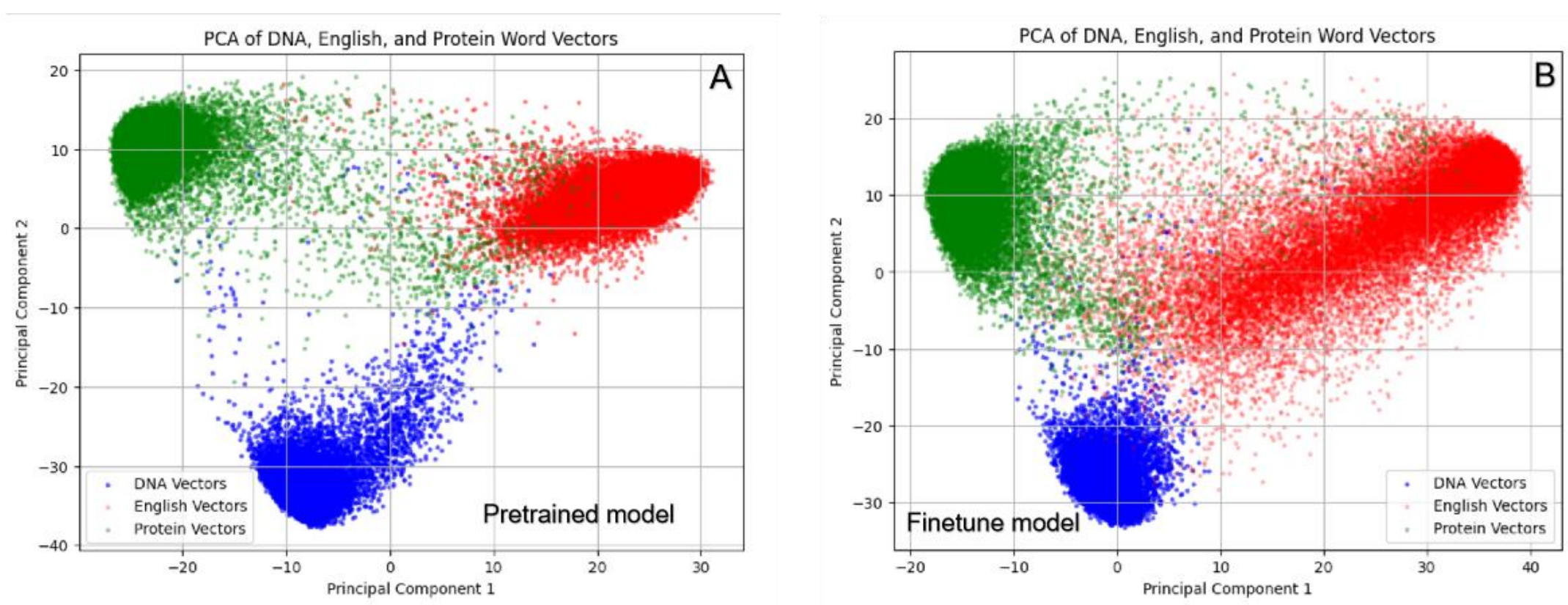

Fig.9. On the left, Figure A shows the word vectors generated by the pre-trained model. In it, blue represents DNA word vectors, red represents English word vectors, and green represents protein word vectors. On the right, Figure B shows the word vectors generated by the fine-tuned model. It can be observed that in Figure A on the left, the three sets of vectors are generally more distant from each other, with clear boundaries. In contrast, in Figure B on the right, the three sets of vectors are closer to one another, and there are overlapping regions. Notably, the overlap between the protein word vector group and the English word vector group is larger than that between the DNA word vector group and the English word vector group. This indirectly suggests why the transfer of English language capabilities to protein sequences is more effective than to DNA sequences.

# 4 Discussion

## 4.1 Facilitating Multilingual Capability Transfer

Based on the experiments conducted, we can summarize several empirical methods for transferring natural language capabilities to DNA sequences:

1 Select Structurally Similar Tasks for Research: Most studies on multilingual capability transfer indicate that structural similarity, rather than semantic similarity, forms the foundation for effective capability transfer. In the XTREME multilingual evaluation benchmark, various tasks are available, but only the text pair similarity task has a structurally similar counterpart in biological sequences. Other tasks, such as the DNA-to-protein coding task, do not have clear analogues. For instance, attempting to transfer from English text similarity tasks to biological sequence tasks yielded poor results. Similarly, transferring from English to Chinese text similarity tasks also performed poorly.

2 Use a Unified Tokenizer, Such as BPE: Utilizing a unified tokenizer like Byte Pair Encoding (BPE) is crucial for enabling capability transfer between languages. This finding aligns with many studies on large multilingual models.

3 Pre-train Large Models on Both Natural Language and Biological Sequence Data: Incorporating a diverse range of natural languages and biological sequences during pre-training can significantly enhance the effectiveness of capability transfer. This mixed data approach helps the model generalize better across different types of sequences.

4 Choose Larger Models if Computational Resources Permit: If computational resources allow, opt for models with larger parameter scales. These models generally offer better performance due to their increased capacity to capture complex patterns.

5 Train Multiple Models with Different Seeds and Select the Best Performing One: The transfer of natural language capabilities to DNA sequence capabilities is highly sensitive to random seeds compared to general tasks. Therefore, it is advisable to train multiple models using different seeds and select the one with the best performance.

## 4.2 New Research Methodologies

The transfer of capabilities from natural language has led to numerous surprising research methods and applications. For instance, pre-training on programming languages has significantly enhanced models' logical reasoning abilities. Similarly, many skills from data-rich languages like

English can be transferred to low-resource languages, leading to a wealth of valuable applications. The transfer of natural language capabilities to biological sequences also holds potential for interesting research methodologies. Below is a feasible approach for such a transfer:

**1. Pre-train Large Models on Both Natural Language and Biological Sequences:**

Develop large models that are pre-trained on both natural language and biological sequence data to capture the nuances of both domains.

**2. Construct a Verifiable Test Dataset for Natural Language to Biological Sequence Transfer:**

Create a dataset specifically designed to test the transferability from natural language to biological sequences. For example, this study used the PAWS-X text similarity dataset as a basis to construct a biological sequence similarity pair dataset.

**3. Fine-tune Models Based on the Above Dataset:**

Fine-tune models using the newly constructed dataset and select the best-performing model as the foundational multimodal model linking natural language to biological sequences.

**4. Fine-tune the Foundational Multimodal Model on Other Natural Language Tasks:**

Further fine-tune the selected "foundational multimodal model" on different types of natural language tasks. For instance, use an English text summarization dataset to fine-tune a new model.

**5. Process Biological Sequences :**

Apply the fine-tuned text summarization model to process long DNA sequences, generating shorter DNA-based summaries. This step explores whether the summarization capability can be effectively transferred from natural language to biological sequences.

**6. Biological Experiment Verification:**

Conduct specific biological experiments to validate the results from step 5. Using text summarization as an example, generate summary DNA sequences from multiple long DNA sequences, then use BLAST (Basic Local Alignment Search Tool) to index and retrieve similar sequences. If the original sequences that produced the summaries are also found to be similar, it indirectly proves that text summarization capability can be transferred from natural language to DNA sequences.

## 5 Conclusion

Language capability transfer in multilingual large models is one of their most attractive features, with structural similarity being key to enabling this capability. This paper builds on relevant research and draws inspiration from the PAWS-X evaluation dataset in XTREME to construct a biosequence similarity judgment evaluation set. This dataset is used to validate the transfer of natural language capabilities to genetic language.

Experimental results show that the multilingual version of the BERT model, after fine-tuning on the English sentence similarity judgment dataset from PAWS-X, can achieve an accuracy of 829% on DNA sequence similarity judgment tasks. In other words, a linguistics expert proficient in multiple natural languages but with no knowledge of biology would still score 89 out of 100 on DNA similarity judgment tests.

Our study is still in its preliminary stages, providing experimental results based on a limited number of evaluation tasks. Further validation of the transfer of natural language capabilities to biological language, along with exploration of corresponding transfer methods and patterns, could enable the powerful reasoning and deduction abilities of large language models to be applied to the niche language of biological sequences. This would undoubtedly introduce a new paradigm to bioinformatics research.

The code and data for this study are open-source and will be continuously updated (24).